\documentclass[11pt, letterpaper, logo, onecolumn, copyright, numbering]{minimax}

\usepackage[authoryear, sort&compress, round]{natbib}
\usepackage{wrapfig}
\usepackage{multirow}
\usepackage{makecell}
\usepackage{subcaption}
\usepackage{cleveref}
\usepackage{xspace}

\usepackage{amsmath,amsfonts,bm}

\def\eqref#1{equation~\ref{#1}}

\def\1{\bm{1}}

\DeclareMathAlphabet{\mathsfit}{\encodingdefault}{\sfdefault}{m}{sl}
\SetMathAlphabet{\mathsfit}{bold}{\encodingdefault}{\sfdefault}{bx}{n}

\def\modelname{VTP}

\newcommand{\eg}{e.g.\xspace}

\let\cite\citep

\title{Towards Scalable Pre-training of Visual Tokenizers for Generation}
\reportnumber{}
\renewcommand{\today}{}

\author[1]{Jingfeng Yao}
\author[2]{Yuda Song}
\author[2]{Yucong Zhou}
\author[1,*]{Xinggang Wang}

\affil[1]{Huazhong University of Science and Technology}
\affil[2]{MiniMax}

\correspondingauthor{xgwang@hust.edu.cn}

\begin{abstract}
The quality of the latent space in visual tokenizers (e.g., VAEs) is crucial for modern generative models. However, the standard reconstruction-based training paradigm produces a latent space that is biased towards low-level information, leading to a foundational flaw: better pixel-level reconstruction accuracy does not lead to higher-quality generation.
This implies that pouring extensive compute into visual tokenizer pre-training translates poorly to improved performance in generation.
We identify this as the ``pre-training scaling problem'' and suggest a necessary shift: to be effective for generation, a latent space must concisely represent high-level semantics.
We present \textit{\modelname}, a unified visual tokenizer pre-training framework, pioneering the joint optimization of image-text contrastive, self-supervised, and reconstruction losses. Our study reveals that \textit{perception-oriented tokenizer pre-training unlocks a new scaling law for generation}, where generative performance scales effectively with compute, parameters, and data allocated to the pre-training of the visual tokenizer.
Our large-scale pre-training experiments demonstrate the following results:
(1) Without modifying DiT training specs and FLOPs, solely scaling VTP pre-training consistently achieves gains in both ImageNet class-conditional and LAION text-to-image generation, while conventional autoencoders stagnate very early at 1/10 of the FLOPs.
(2) VTP achieves 0.36 rFID while simultaneously delivering 78.2\% zero-shot accuracy and 85.7\% linear probing accuracy, surpassing prior unified tokenizers such as VILA-U and UniTok.
(3) Furthermore, the VTP-based diffusion model exhibits exceptionally fast convergence---reaching 2.03 gFID in only 80 epochs without guidance tricks, outperforming previous methods like VA-VAE and RAE---and ultimately scales to achieve a remarkable \textit{1.11 gFID} on ImageNet $256\times256$ generation.
Our code and models are publicly available at \href{https://github.com/MiniMax-AI/VTP}{https://github.com/MiniMax-AI/VTP}.
\end{abstract}

\begin{document}

\maketitle

\section{Introduction}

Latent Diffusion Models (LDMs)~\cite{ldm} employ a visual tokenizer, such as a VAE~\cite{vae}, to compress visual signals into a latent space. Typically, visual tokenizers are pre-trained in a separate stage using a reconstruction objective.

\begin{figure}[t]
    \centering
    \includegraphics[width=0.95\linewidth]{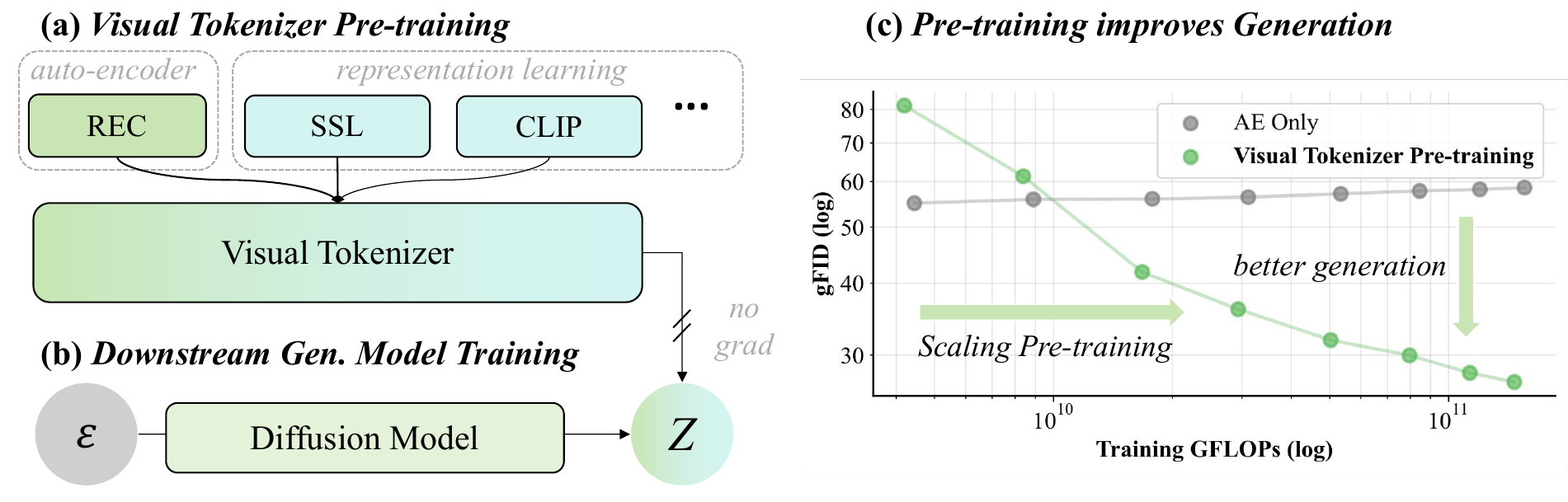}
    \caption{\textbf{Visual Tokenizer Pre-training.} We revisit the visual tokenizer pre-training in LDM~\cite{ldm} from a representation learning perspective. Critically, while keeping the diffusion model (e.g., DiT~\cite{dit}) training configuration and FLOPs fixed, our method improves generation solely by scaling the tokenizer's pre-training to learn a better-structured latent space. }
    \label{fig:first}
\end{figure}

However, a clear paradox has emerged: better reconstruction does not guarantee better generation. Instead, a noticeable trade-off between the two objectives is widely observed~\cite{sd3, sana, vavae}. 
It implies that scaling the computational investment in pre-training, while potentially further improving reconstruction performance, carries the risk of compromising generation performance (see \cref{fig:first} (c)), which is consistent with prior work~\cite{scale_vae}.
We note that this limitation of reconstruction-only training may arise because the objective biases the latent space toward low-level information and, as training scales up, increasingly drives it away from the structured latent space we ultimately desire, thereby motivating the search for tokenizer pre-training schemes that genuinely scale—a challenge we term the \textit{``pre-training scaling problem''}.

Unlike conventional approaches that emphasize low-level information, we propose that an effective latent space for generation should efficiently encode the core visual semantics. Early explorations have already demonstrated the value of this principle through two primary pathways. Some works explicitly enrich the latent space with specific semantic objectives—for instance, by concatenating optical flow for motion or leveraging powerful pre-trained features to the latent space, like VideoJAM~\cite{videojam} and ReDi~\cite{redi}. Others implicitly structure the space using semantic constraints, an approach seen in methods like VA-VAE~\cite{vavae} and REPA-E~\cite{repa-e}, which regularize the VAE's feature space with representational priors.
While promising, these efforts remain preliminary and do not explore broader scaling properties. 

\begin{wrapfigure}{r}{0.42\textwidth}
    \centering
    \vspace{-12pt}
    \includegraphics[width=0.40\textwidth]{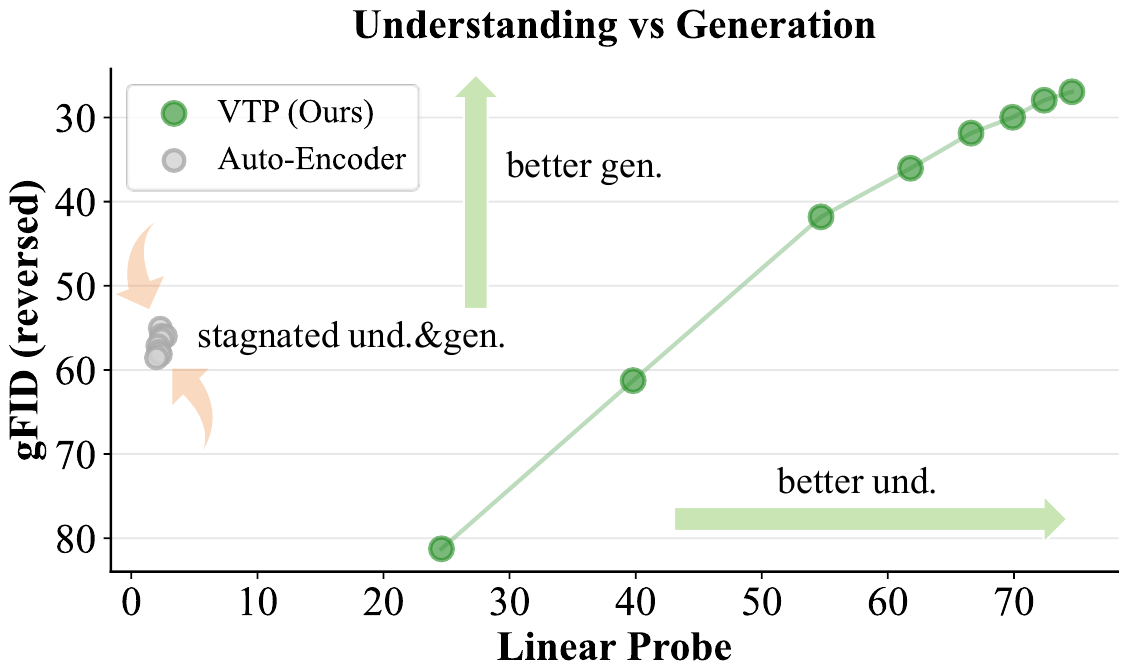}
    \caption{\textbf{Understanding is a key driver of generation.} We observe a strong positive correlation between the comprehension and generative capabilities of the latent space during visual tokenizer pre-training.}
    \label{fig:und_vs_gen}
    \vspace{-24pt}
\end{wrapfigure}
To address this challenge, we present \textbf{\modelname}, a novel pre-training framework for visual tokenizers. The core contribution of our work is a redesigned, scalable paradigm for visual tokenizer pre-training that benefits generation. This is achieved by jointly optimizing the model across a spectrum of visual representation tasks, including cross-modal alignment, global semantic understanding, local spatial perception, and low-level pixel reconstruction. Technically, our framework is built upon a vision transformer (ViT)~\cite{vit} based Auto-Encoder.
Building on the flexibility of the ViT architecture for representation learning, we integrate a suite of diverse learning objectives. First, cross-modal image-text contrastive learning is employed to instill a global semantic understanding~\cite{clip, openclip}. This is complemented by integrating established self-supervised learning techniques, notably self-distillation and mask image modeling~\cite{mae, ibot, dino, dinov2}, to enhance the model's spatial-semantic perception. Throughout this multi-task learning process, the pixel-level reconstruction objective is consistently applied to preserve fine-grained visual details for generation. We posit that this holistic training paradigm encourages the latent space to form a unified and rich representation of visual information, which is instrumental in boosting the fidelity and semantic coherence of the generated outputs. (\cref{sec:method})

Through extensive experiments, we reveal that \textit{perception-oriented tokenizer pre-training unlocks a new scaling frontier for generation}:
(1) Understanding is a key driver of generation: The introduction of semantic understanding and perception tasks enhances the generative capability of models initially pre-trained solely on reconstruction. We observe a strong positive correlation between the semantic quality of the latent space and its generative performance. While these tasks differ in paradigm, they consistently inject more meaningful representations, leading to significant gains in downstream generation. (see \cref{fig:und_vs_gen})
(2) Superior Scalability for Generation: \modelname~is the first visual tokenizer to demonstrate scaling properties. Its generative performance improves steadily as we scale up training compute (FLOPs), model parameters, and dataset size of the visual tokenizer. This stands in stark contrast to traditional tokenizers pre-trained only on reconstruction, whose performance rapidly saturates and shows negligible gains with increased scale.

We conduct extensive experiments on ImageNet~\cite{imagenet} class-conditional generation and LAION~\cite{laion400m} text-to-image generation. (1) We demonstrate new scalability for visual tokenizer pre-training: increasing pre-training compute accelerates the convergence of downstream VTP-based diffusion models. Specifically, scaling compute by 10$\times$ yields a 65.8\% FID improvement for DiT on ImageNet, overcoming the early saturation of conventional autoencoders. (2) A systematic analysis reveals that diverse perception losses consistently improve general generation quality (\cref{sec:exp}), and (3) specific losses offer targeted benefits for downstream tasks; \eg, CLIP loss significantly enhances text-to-image generation (\cref{sec:t2i}).
(4) Our final model gets a strong performance, achieving 0.36 rFID on ImageNet alongside 78.2\% zero-shot and 85.7\% linear probing accuracy, outperforming unified tokenizers like VILA-U~\cite{vila-u} and UniTok~\cite{unitok}. Coupled with a DiT~\cite{dit,vavae}, it enables exceptionally fast convergence (2.03 FID in just 80 epochs, unguided), surpassing prior approaches such as VA-VAE~\cite{vavae} and RAE~\cite{rae}, and ultimately reaches a remarkable 1.11 gFID on generation tasks.

To sum up, our contribution could be summarized as follows:
\begin{itemize}
\item We propose Visual Tokenizer Pre-training (VTP), a framework integrating contrastive, self-supervised, and reconstruction objectives to build a perception-oriented tokenizer pre-training for generative models.
\item We demonstrate a new scaling property for visual tokenizers: overcoming the early saturation of reconstruction-only training, downstream generation quality consistently improves as VTP pre-training scales in compute, parameters, and data.
\item VTP pushes the boundaries of unified understanding and generation of visual encoder, achieving an impressive 1.11 gFID on ImageNet generation alongside 0.36 rFID, 78.2\% zero-shot, and 85.7\% linear probing accuracy.
\end{itemize}
\section{Related Work}

\subsection{Pre-training and Representation Learning}

Pretraining is typically a scalable paradigm for boosting downstream task performance by first optimizing models on large-scale data with specific objectives. 
The early paradigm relied on supervised pretraining—such as ImageNet classification~\cite{resnet, vit}—and transferred weights to downstream tasks like detection~\cite{faster-rcnn} and segmentation~\cite{upernet}. A recent paradigm shift has focused on weakly-supervised and unsupervised methods to enable pretraining at larger scales. For instance, CLIP~\cite{clip} uses image-text contrastive learning with weak supervision by minimizing the distance between image and text features. SigLIP~\cite{siglip} further optimizes this process via a sigmoid loss for large-scale training.
Another branch, self-supervised learning (SSL), learns directly from unlabeled data. Methods like MAE~\cite{mae} and BEiT~\cite{beit} adopt masked image modeling (MIM), training models to reconstruct masked patches. DINO~\cite{dino} utilizes self-distillation to enforce multi-view classification consistency. iBOT~\cite{ibot} and DINOv2~\cite{dinov2} combine MIM with self-distillation to learn more generalized representations. 

Despite these advances, within the explicitly decoupled, two-stage framework of LDMs~\cite{ldm}—comprising a visual tokenizer followed by a generative model—how to pretrain the first-stage tokenizer to enhance second-stage generative performance has not been systematically explored.

\subsection{Latents with Pre-trained Representations}

Previous work has explored the use of visual representations to structure the latent space, which falls into two categories. The first employs a distillation objective: VA-VAE~\cite{vavae} aligns its latent space with features of visual foundation models to alleviate the trade-off between reconstruction and generation. ImageFolder~\cite{imagefolder} decouples semantic and pixel-level feature spaces to improve autoregressive generation. MAETok~\cite{maetok} enhances latent representations by incorporating DINOv2 features into its MIM pre-training objective. REPA-E~\cite{repa-e}concurrently optimizes the feature space of the VAE during DiT training by leveraging supervision from a pre-trained foundation model. I-DeTok~\cite{l-detok} improves the latent space's suitability for both autoregressive and diffusion models generation via a joint pre-training strategy that employs MIM and noise injection. The second strand directly utilizes pre-trained representations for generation. For instance, BLIP3-o~\cite{blip3o} regresses SigLIP features and employs an SD-XL-based~\cite{sdxl} decoder to boost efficiency. Recently, RAE~\cite{rae} leverages DINOv2 features and trains a separate pixel decoder for reconstruction. 

However, these methods are inherently limited by existing foundational models, often leading to a low performance ceiling or substantial reconstruction loss. Concurrently, while previous studies achieved performance improvements under specific configurations, the scalability of the proposed methods generally remains unverified.

\begin{figure*}[t]
    \centering
    \includegraphics[width=\linewidth]{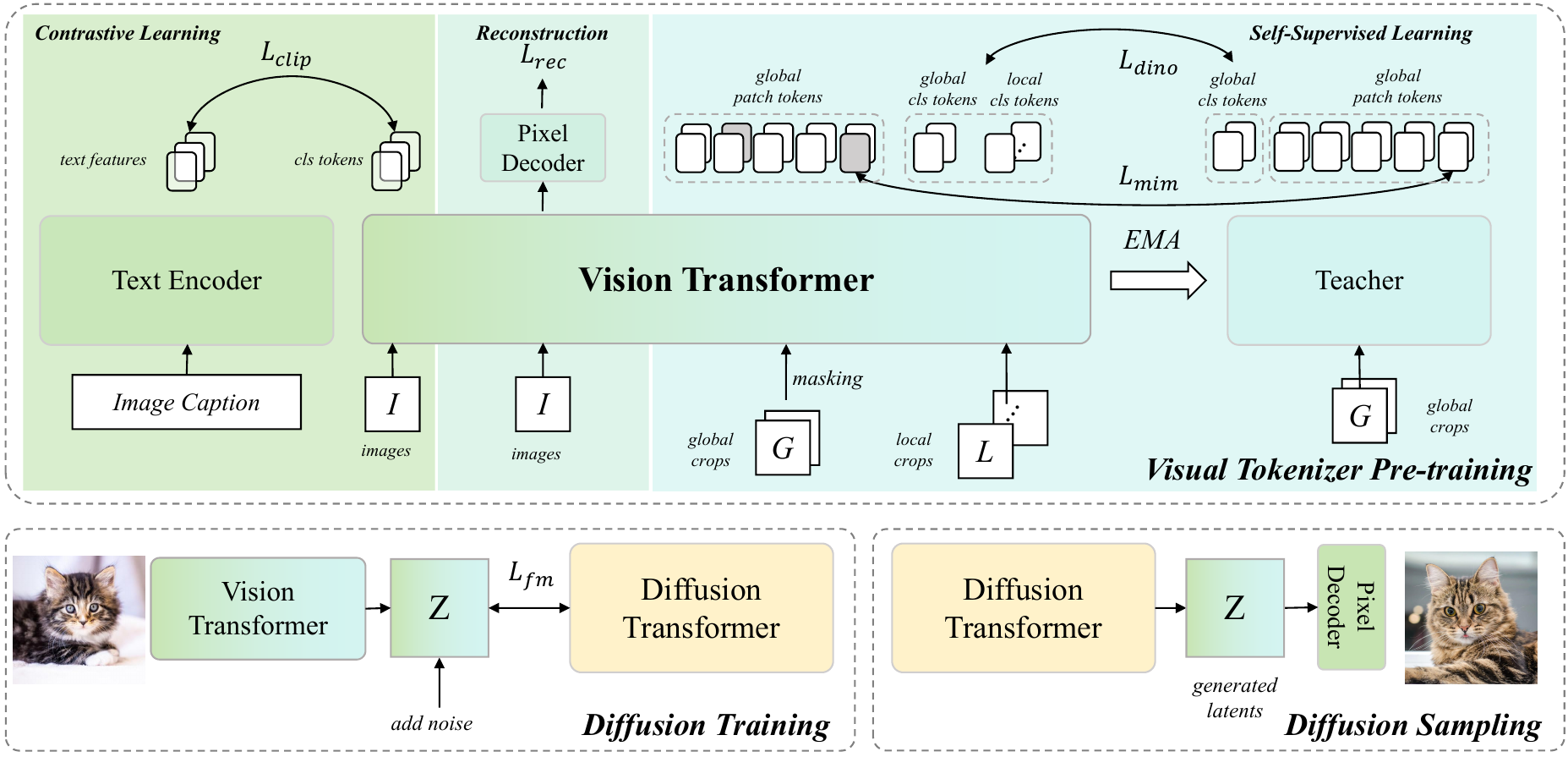}
    \caption{\textbf{Overwiew of Visual Tokenizer Pre-training (\modelname).} By integrating representation learning (image-text contrastive~\cite{clip} and self-supervised learning~\cite{dinov2}) with reconstruction within a Vision Transformer Auto-Encoder, we find that \modelname~exhibits a \textbf{well-behaved scaling property} for generative performance.}
    \label{fig: overview}
\end{figure*}
\section{Visual Tokenizer Pre-training}
\label{sec:method}
Our work introduces a scalable visual tokenizer pre-training paradigm that benefits generation. To this end, we integrate representation learning objectives with the conventional reconstruction loss to learn visual representations that are semantically rich, accurate in reconstruction, and generation-friendly (see \cref{fig: overview}). 

\subsection{Architecture}
\label{sec:vitae_arch}

Leveraging its flexibility in learning visual representations, our visual tokenizer uses a fully Vision Transformer (ViT) architecture. In line with standard autoencoder designs, we introduce a bottleneck that maps visual information into a $d$-dimensional latent space. Encoder features are leveraged by the text encoder, EMA teacher, and pixel decoder to facilitate their distinct training objectives.

\subsection{Visual Reconstruction}
\label{sec:rec}

Given an image $I \in \mathbb{R}^{3 \times H \times W}$, we compress it into a latent space $\mathbb{R}^{d \times H/16 \times W/16}$ using a visual tokenizer and subsequently reconstruct into $I'$ via a pixel decoder, which lifts latents back to the feature space, refines them with N ViT blocks, and reconstructs images in pixel space through a final pixel-shuffle layer.

The reconstruction task is challenged by the poor compatibility of GAN loss~\cite{vqgan} with the ViT architecture, which causes large gradient norms and low training stability. 
To address this, we employ a two-stage training strategy. In the first stage (i.e., the pre-training stage), all parameters are jointly optimized by minimizing a composite loss function comprising the $\mathcal{L}_1$ loss and a perceptual loss $\mathcal{L}_{\text{perceptual}}$~\cite{lpips} between $I$ and $I'$. 
During the second stage, the visual tokenizer remains frozen while the pixel decoder is fine-tuned with a GAN objective to improve fidelity.

The overall reconstruction loss $\mathcal{L}_{\text{rec}}$ during pre-training is defined as:
\begin{equation}
\mathcal{L}_{\text{rec}} = \mathcal{L}_1 + \mathcal{L}_{\text{perceptual}}
\end{equation}

\subsection{Self-Supervised Learning}
\label{sec:method_ssl}

Following DINOv2~\cite{dinov2}, our self-supervised learning framework comprises two components: masked image modeling (MIM)~\cite{mae, ibot} and self-distillation~\cite{dino}.

For a given image $I$, we apply data augmentation to obtain global and local views $I_\text{global}$ and $I_\text{local}$. In MIM, adhering to~\cite{ibot}, $I_\text{global}$ is patch-embedded and fed directly to an EMA teacher, while its masked version is processed by the visual tokenizer, optimizing the complementary masking loss $\mathcal{L}_\text{mim}$. For self-distillation, similar to~\cite{dino}, $I_\text{global}$ and $I_\text{local}$ are passed to the visual tokenizer, and $I_\text{global}$ to the EMA teacher, with the cross-entropy loss $\mathcal{L}_\text{dino}$ applied to their pseudo-label predictions.

Therefore, the overall self-supervised learning loss is defined as:
\begin{equation}
\mathcal{L}_{ssl} = \mathcal{L}_{mim} + \mathcal{L}_{dino}
\end{equation}

\subsection{Contrastive Learning}
\label{sec:method_clip}

Given a batch of image-text pairs, we encode the image $I$ and text $T$ using a visual tokenizer and a text encoder, respectively, to obtain their visual and textual features. Following CLIP, we then maximize the similarity of the corresponding (positive) image-text pairs while minimizing the similarity of the remained non-corresponding (negative) pairs. This objective is formulated as the contrastive loss $\mathcal{L}_\text{clip}$.

\subsection{Overall Objective}

Building upon the preceding components, we integrate them into a unified pre-training framework. The overall training objective for our visual tokenizer pre-training is formulated as a weighted combination of the aforementioned losses:

\begin{equation}
\mathcal{L}_{\text{total}} = \lambda_{\text{rec}}\mathcal{L}_{\text{rec}} + \lambda_{\text{ssl}}\mathcal{L}_{\text{ssl}} + \lambda_{\text{clip}}\mathcal{L}_{\text{clip}}
\end{equation}
where $\lambda_{\text{rec}} > 0$, $\lambda_{\text{ssl}} \geq 0$, and $\lambda_{\text{clip}} \geq 0$ are balancing coefficients that control the contribution of each objective. This multi-task learning scheme enables the model to concurrently develop high-fidelity reconstruction capability, semantically rich representation learning, and cross-modal alignment, thereby establishing a robust and scalable visual tokenizer for diverse generation tasks.

\subsection{Batch Sampling}

We observe a significant disparity in optimal batch sizes across different training paradigms. Contrastive learning frameworks like CLIP demand extremely large batches (\eg, 16k or 32k), while self-supervised and reconstruction objectives are typically effective with orders of much smaller batches (\eg, 4k).

Given an input batch of \( B \) image-caption pairs, all samples are used for CLIP training, e.g. \(  B_{\text{clip}} = B \). \( B_{\text{ssl}} \) and \( B_{\text{rec}} \) are random sampled from \( B \) to accommodate the divergent batch size requirements of self-supervised learning and reconstruction.
\section{Experiments}
\label{sec:exp}

\subsection{Implementation Details}

\paragraph{Pre-training}
Our model architecture builds upon the Vision Transformer (ViT) implemented in~\cite{dinov3}. We incorporate QKNorm~\cite{qknorm} to enhance training stability. We employ a 12-layer transformer with a hidden dimension of 768 as the text encoder, and a 4-layer ViT-Large layer as the pixel decoder for fast experimentation. In designing the latent bottleneck, we primarily adopt a dimension of 64, following~\cite{unitok}, to balance semantic comprehension with reconstruction quality. An ablation study on this configuration is conducted by varying the dimension to 256. 
We use an internally filtered version of DataComp-1B~\cite{imagenet} with 277M samples for tokenizer pretraining, and ImageNet~\cite{imagenet} for downstream DiT training.
We set $B_{\text{clip}}=16k$, $B_{\text{ssl}}=4k$ and $B_{\text{rec}}=2k$. For weighting, we set $\lambda_{\text{rec}} = 0.1$, while $\lambda_{\text{clip}}$ and $\lambda_{\text{ssl}}$ are set to either 0 or 1. We find that a smaller reconstruction weight contributes to improved generative performance.
For self-supervised and contrastive pretraining implementation, we closely follow the established practices of DINOv2~\cite{dinov2} and OpenCLIP~\cite{openclip}.

\paragraph{Downstream DiT training \& evaluation}
We train the Diffusion Transformer (DiT)~\cite{dit} under a fixed configuration to evaluate the generative capability of our visual tokenizer. Specifically, we follow LightningDiT~\cite{vavae} as a strong baseline. We report FID-10k scores obtained with a LightningDiT-B~\cite{vavae} model trained on ImageNet~\cite{imagenet} for 80 epochs under a consistent protocol. 
For the reconstruction evaluation, the performance of all tokenizers is assessed on the standard ImageNet validation set at a resolution of 256. For most cases, we report rFID as the reconstruction metric.
For understanding evaluation,
we evaluate representation performance on ImageNet using linear probing. We do not employ the feature enhancements common in the DINO~\cite{dinov2, dinov3} series, which can substantially increase linear probing scores by leveraging multi-layer features. Instead, we probe only the reduced-dimensionality features from the bottleneck, thereby directly evaluating the inherent properties of the latent features, and report ImageNet Top-1 acc. as the understanding metric.

\subsection{Auto-Encoder with Vision Transformers}
\label{sec:vitae_exp}

\begin{figure}[t]
    \centering
    \begin{minipage}[b]{0.32\textwidth}
        \centering
        \resizebox{\textwidth}{!}{%
        \begin{tabular}{l|ll|cc}
        \toprule
        \textbf{Arch.} & \textbf{FLOPs} & \textbf{\#Params} & \textbf{\textcolor{gray}{\textit{r}}PSNR} & \textbf{\textcolor{gray}{\textit{g}}FID} \\
        \midrule
        CNN~\cite{ldm}  &  389.4G &   70.3M & 30.63 &  59.53 \\
        \midrule
        ViT-B~\cite{vit}  &  87.7G & 171.2M & 30.72 & 58.40 \\
        ViT-L~\cite{vit} & 311.1G  & 607.2M & \textbf{31.28} &  \textbf{53.51} \\
        \bottomrule
        \end{tabular}}
        \captionof{table}{\textbf{AutoEncoder Performance with Different Architectures.} ViT serves as a comparable alternative to CNNs, enabling simpler pre-training scaling experiments.}
        \label{tab:vit_baseline}
    \end{minipage}
    \hfill
    \begin{minipage}[b]{0.65\textwidth}
        \centering
        \includegraphics[width=\textwidth]{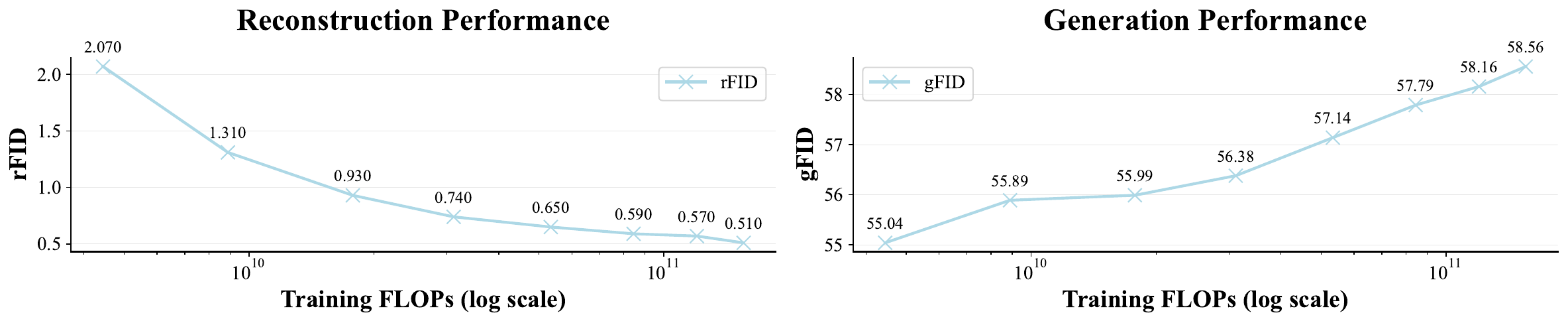}
        \captionof{figure}{\textbf{Reconstruction Only Training Target CANNOT Lead to Effective Scaling for Downstream Diffusion Models.} As training progresses, the tokenizer's reconstruction performance improves, while its generative performance degrades concurrently. It reveals the \textit{inadequacy of pure reconstruction tasks for scalable tokenizer pre-training.}}
        \label{fig:ae_only}
    \end{minipage}
\end{figure}

We begin by demonstrating that a Vision Transformer (ViT) can serve as an effective substitute for CNNs in standard reconstruction tasks.
We construct a ViT visual tokenizer with a symmetric encoder and decoder. It has the specification of f16d64, where `$f$' denotes downsample ratio (or patch size for ViT) and `$d$' denotes bottleneck dimension. A two-stage training pipeline discussed in \cref{sec:rec} is adopted to enhance training stability.
Then, we implement the convolutional LDM architecture~\cite{ldm} under the same specifications. We use ImageNet at 256 resolution for training and testing.

As illustrated in \cref{tab:vit_baseline}, we evaluate their reconstruction and generation performance, observing that ViT-L achieves a reconstruction PSNR of 31.28 and a gFID of 53.51, on par with LDM. While it utilizes more parameters, it requires lower computational cost. These findings are consistent with previous observations~\cite{scale_vae, magi-1}, suggesting that the simple design of this ViT tokenizer architecture is effective.

\subsection{Scaling up Visual Tokenizer Pre-training}
\label{sec:scaling_flops}

\begin{figure*}[t]
    \centering
    \begin{subfigure}[b]{1.0\textwidth}
        \centering
        \includegraphics[width=\textwidth]{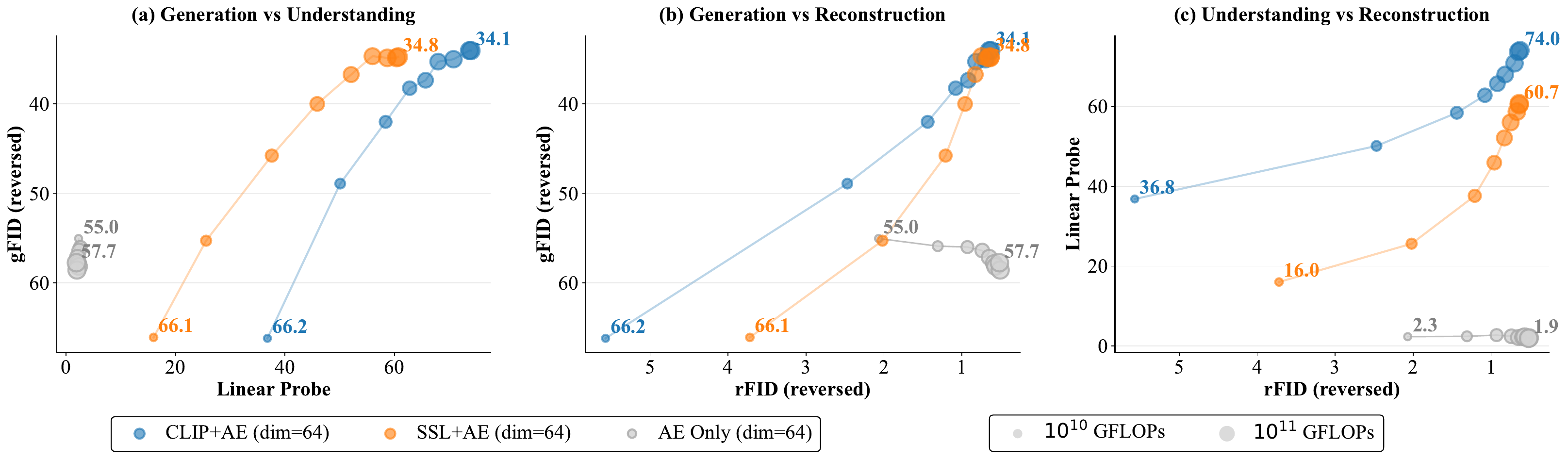}
        \label{fig:clip_ae_dim64}
    \end{subfigure}
    \begin{subfigure}[b]{1.0\textwidth}
        \centering
        \includegraphics[width=\textwidth]{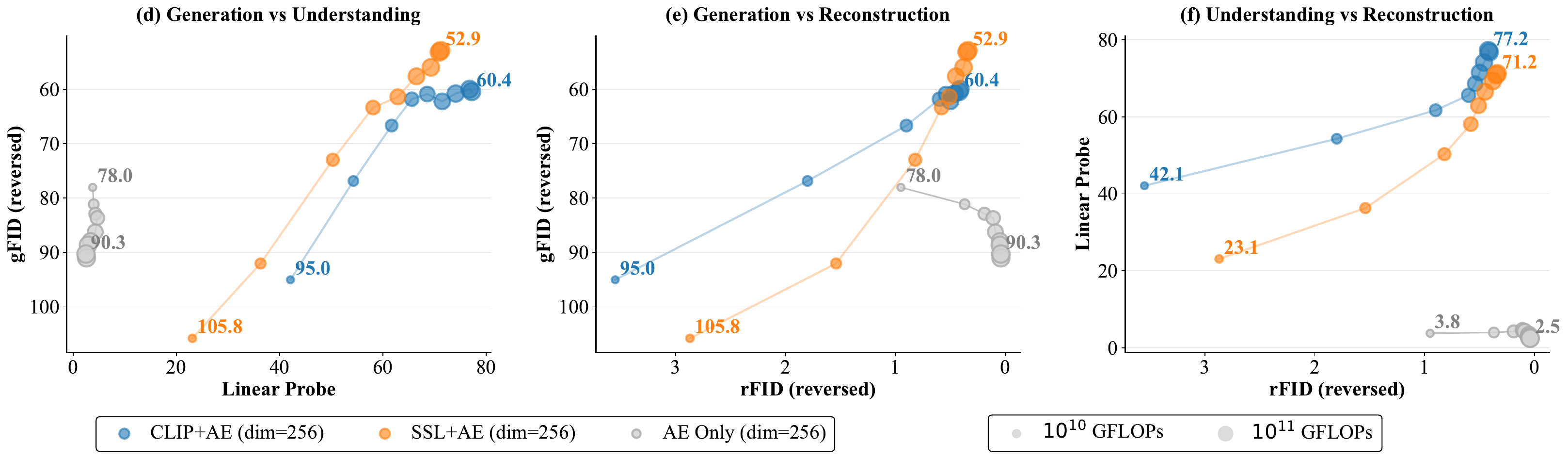}
        \label{fig:clip_ae_dim256}
    \end{subfigure}
    \caption{
        \textbf{Scalability of CLIP+AE \& SSL+AE Visual Tokenizer Pre-training.} Scaling properties under different strategies and bottleneck dimensions. Our method shows correlated growth in generation and comprehension with compute, while VAE-based tokenizer performance rapidly saturates.
    }
    \label{fig:unified_scaling_analysis}
\end{figure*}

Our work focuses on how to scale up the tokenizer pre-training to improve the model's capabilities for downstream generative training.

We conduct three distinct scaling-up experiments. For these experiments, we employ a ViT-L backbone as the encoder and a lightweight decoder composed of 4 ViT-L layers to facilitate rapid training and inference. All models are trained on the 277M DataComp-filtered dataset introduced above.

\paragraph{Scaling with reconstruction only CANNOT help generation.}
\label{sec:scaling_rec}

Initially, we scale up the training computation for a standard reconstruction tokenizer.
As illustrated in \cref{fig:ae_only}, we observe a scaling paradox: the model's reconstruction performance improves substantially with increased training compute, with rFID improving from 2.0 to 0.5. However, its generative performance in fact slightly degrades, as indicated by the gFID rising from 55.04 to 58.56. We posit that this phenomenon occurs because the reconstruction objective effectively guides the model to capture low-level details but provides insufficient incentive for learning high-level semantic representations, which are crucial for generation. Reconstruction task itself does not exhibit scalability in pretraining for downstream generation.

\paragraph{Scaling with different understanding tasks helps generation in a similar way.} Then, we scale up visual tokenizer pre-training with the assistance of representation tasks.
As described in \cref{sec:method}, we integrate the reconstruction task with either image-text contrastive learning (CLIP)~\cite{clip} or self-supervised learning (SSL, specifically DINOv2)~\cite{dinov2} in a joint training framework. These two hybrid approaches are denoted as CLIP+AE and SSL+AE, respectively. To further substantiate the robustness of our conclusions, we include an additional experimental configuration with a latent dimension of $d=256$. For a fair comparison, all Autoencoders (AEs) across different latent dimensions were trained under an identical computational budget. We concurrently monitor four key metrics: understanding performance, reconstruction fidelity, generation quality, and training FLOPs for a comprehensive evaluation. 

Our experimental results, summarized in \cref{fig:unified_scaling_analysis}, lead to the following key observations:

(1) \textit{Feasibility of Hybrid Objectives}: Hybrid training combining representation learning with reconstruction is viable. As evidenced by the \cref{fig:unified_scaling_analysis} (c)\&(f), traditional autoencoders (AEs) trained solely on reconstruction maintain low understanding performance. In contrast, when augmented with representation learning objectives—either CLIP or SSL—both understanding and reconstruction metrics exhibit stable, simultaneous improvement.

(2) \textit{Negative Impact of Pure Reconstruction}: Solely relying on reconstruction proves counterproductive for downstream generation tasks. The \cref{fig:unified_scaling_analysis} (b)\&(e) illustrates a negative yield in reconstruction-only AEs: as the computational budget increases, reconstruction performance improves but the generation performance degrades.

\begin{figure*}[t]
    \centering
    \includegraphics[width=\textwidth]{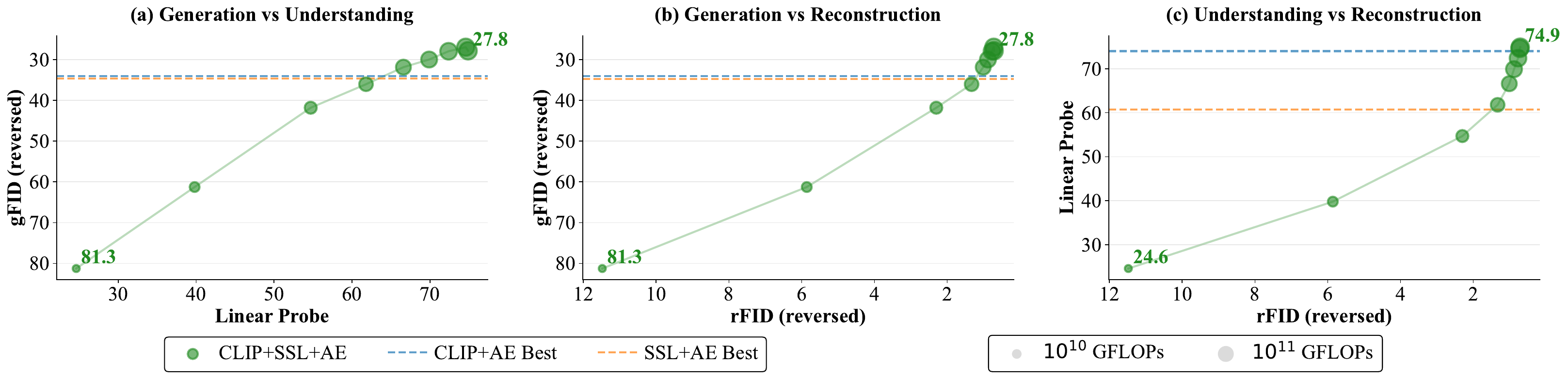}
    \caption{\textbf{Scalability of CLIP+SSL+AE Visual Tokenizer Pre-training.} Under the same computational budget, the f16d64 tokenizer trained with joint CLIP and SSL representation learning achieves the best performance in both generation and comprehension.}
    \label{fig:3in1}
\end{figure*}

\begin{figure*}[t]
    \centering
    \begin{subfigure}[b]{0.32\textwidth}
        \centering
        \includegraphics[width=\textwidth]{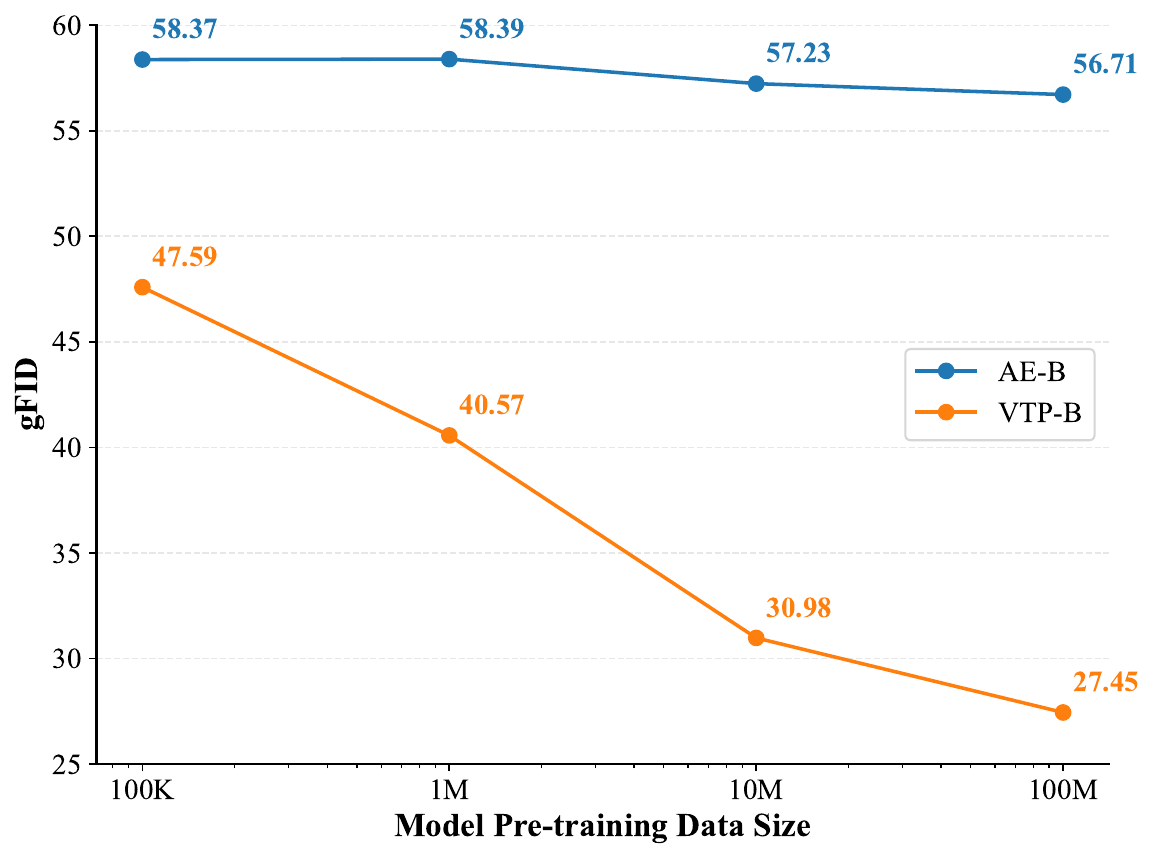}
        \caption{\textbf{Data Scaling.}}
        \label{fig:data_scaling}
    \end{subfigure}
    \hfill
    \begin{subfigure}[b]{0.32\textwidth}
        \centering
        \includegraphics[width=\textwidth]{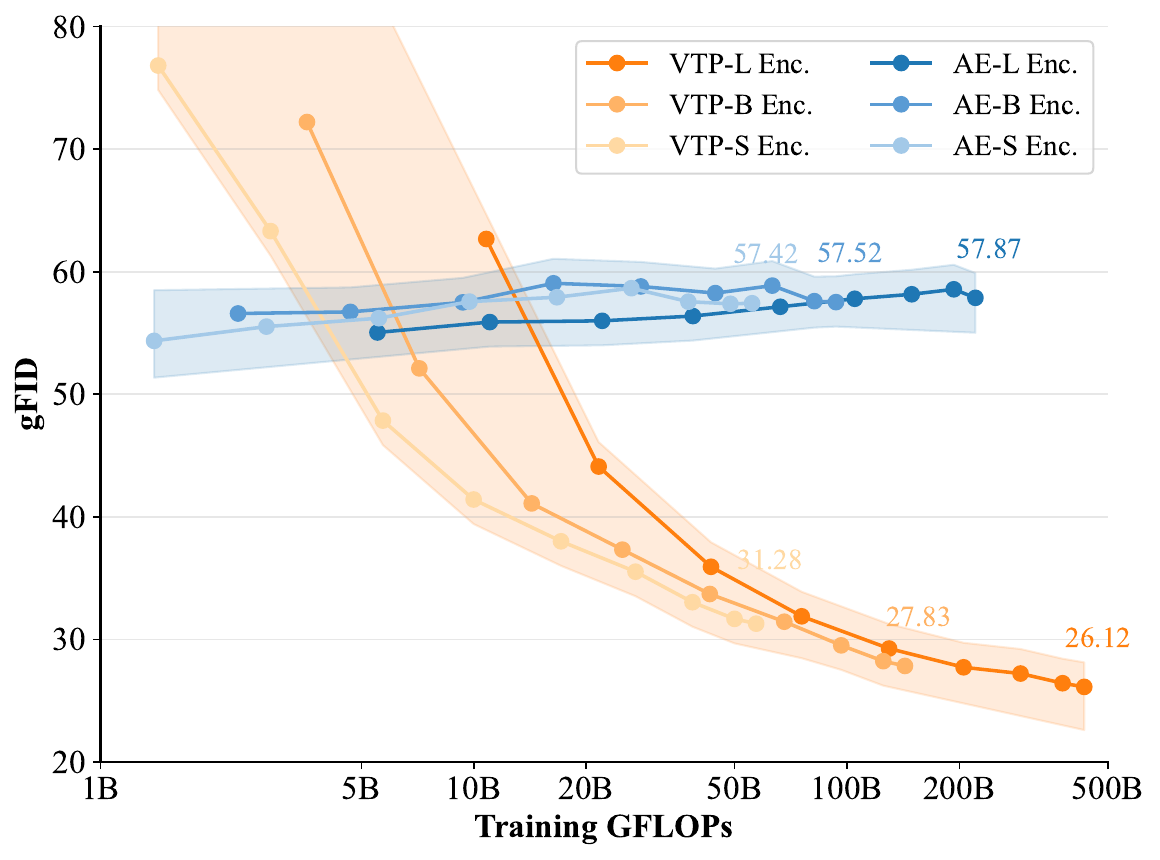}
        \caption{\textbf{Encoder Scaling.}}
        \label{fig:encoder_scaling}
    \end{subfigure}
    \hfill
    \begin{subfigure}[b]{0.32\textwidth}
        \centering
        \includegraphics[width=\textwidth]{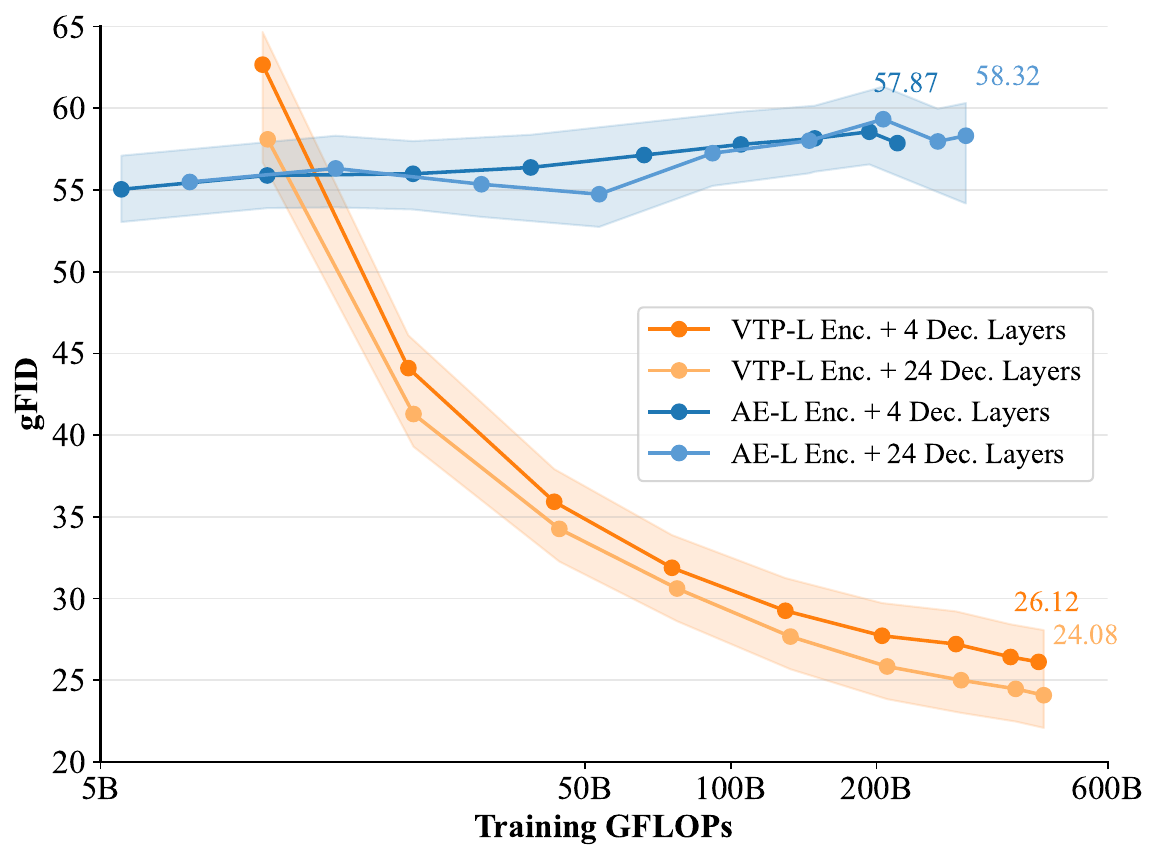}
        \caption{\textbf{Decoder Scaling.}}
        \label{fig:generation_comparison}
    \end{subfigure}
    \caption{
    \textbf{Scalability of data and parameters.}
        We observe a new scaling property: the DiT generation performance within fixed training FLOPs increases while its tokenizer has larger model sizes and training data.
        }
    \label{fig:scaling_results}
\end{figure*}

(3) \textit{Understanding as the Key Driver}: The integration of semantic understanding tasks counteracts this negative effect and emerges as the dominant factor for improving generation. The reversal of this trend is visible in the \cref{fig:unified_scaling_analysis} (a)\&(d) and (b)\&(e). Specifically, our visual tokenizer pre-training, which jointly optimizes for reconstruction and representation learning, enables continuous, concurrent improvement in reconstruction, understanding, and generation as pre-training scales. Conversely, focusing exclusively on reconstruction optimization yields superior reconstruction, but leads to the stagnation of both understanding and generation performance. These observations imply that understanding is the key driving force necessary for effective generation.

(4) \textit{Generality of Representation Learning}: Diverse representation learning paradigms, including CLIP and SSL, consistently enhance generation performance. Despite significant differences in their training frameworks, they share a critical mechanism: enriching the semantic understanding within the latent space. Although their scaling behaviors differ slightly, both methods substantially improve the efficacy of visual tokenizer pre-training for downstream generative tasks. This observation also suggests that new and emerging representation learning techniques can be seamlessly integrated to establish even better performance bounds.

\paragraph{Scaling with multiple understanding tasks gets better performance.}

Subsequently, we introduce, to the best of our knowledge, the first integration of contrastive, self-supervised, and reconstruction objectives (CLIP+SSL+AE) for visual tokenizer pre-training. Our experiments demonstrate that this training paradigm is feasible and stable. This multi-objective framework enables the tokenizer to capture multi-scale features, enhancing both semantic alignment and spatial fidelity. 
As shown in \cref{fig:3in1}, our method under the f16d64 setting achieves a higher generative upper bound (gFID=27.8) alongside better understanding performance (74.9\% linear probing accuracy) under a fixed computational budget. All subsequent data and parameter scaling experiments are based on this pre-training configuration.

\subsection{Scaling Properties with Parameters}
\label{sec:scaling_params}

\modelname~demonstrates a new interesting parameter scalability, as its downstream DiT generative performance improves consistently with increased previous tokenizer model size.

We first investigate encoder scaling by training three ViTs of varying sizes using CLIP+SSL+AE and a baseline AE. 
As shown in \cref{fig:scaling_results} (b), the generative performance of the AE remains stagnant at about 57, regardless of the model capacity (from 20M to 300M parameters). In contrast, \modelname~exhibits a clear scaling trend: its gFID improves steadily from 31.28 to 26.12 as the model size grows, forming a well-defined parameter scaling curve. 
We then proceed to scale up the pixel decoder. Our findings indicate that this architectural expansion also leads to a correlated improvement in generative performance, with the gFID score decreasing from 26.12 to 24.08. 

\subsection{Scaling Properties with Data}
\label{sec:scaling_data}
The scale of training data is also crucial for the generalization ability of the tokenizer. To validate this, we constructed four subsets of varying scales—100K, 1M, 10M, and 100M—by randomly sampling from the Datacomp-1B dataset. We trained both VTP-ViT-Large and AE-ViT-Large architectures on these subsets for 1.1 billion samples each and evaluated their generation performance. The results are summarized in \cref{fig:scaling_results} (a).

We draw the following observations from the results: First, VTP consistently outperforms the conventional autoencoder across all data scales. More importantly, the generative performance of the autoencoder shows negligible improvement with increased data, with its FID score merely decreasing from 58.37 to 56.71. In stark contrast, the performance of VTP improves significantly as the volume of training data grows, with its FID score substantially improving from 47.59 to 27.45. Notably, the downstream DiT training FLOPs remained strictly identical. This compellingly demonstrates that the introduced representation learning effectively enhances the data scalability of the visual tokenizer.

\section{Scaling to Text-to-Image Generation}
\label{sec:t2i}

\begin{figure}[t]
    \centering
    \begin{subfigure}[b]{0.64\textwidth}
        \centering
        \includegraphics[width=\textwidth]{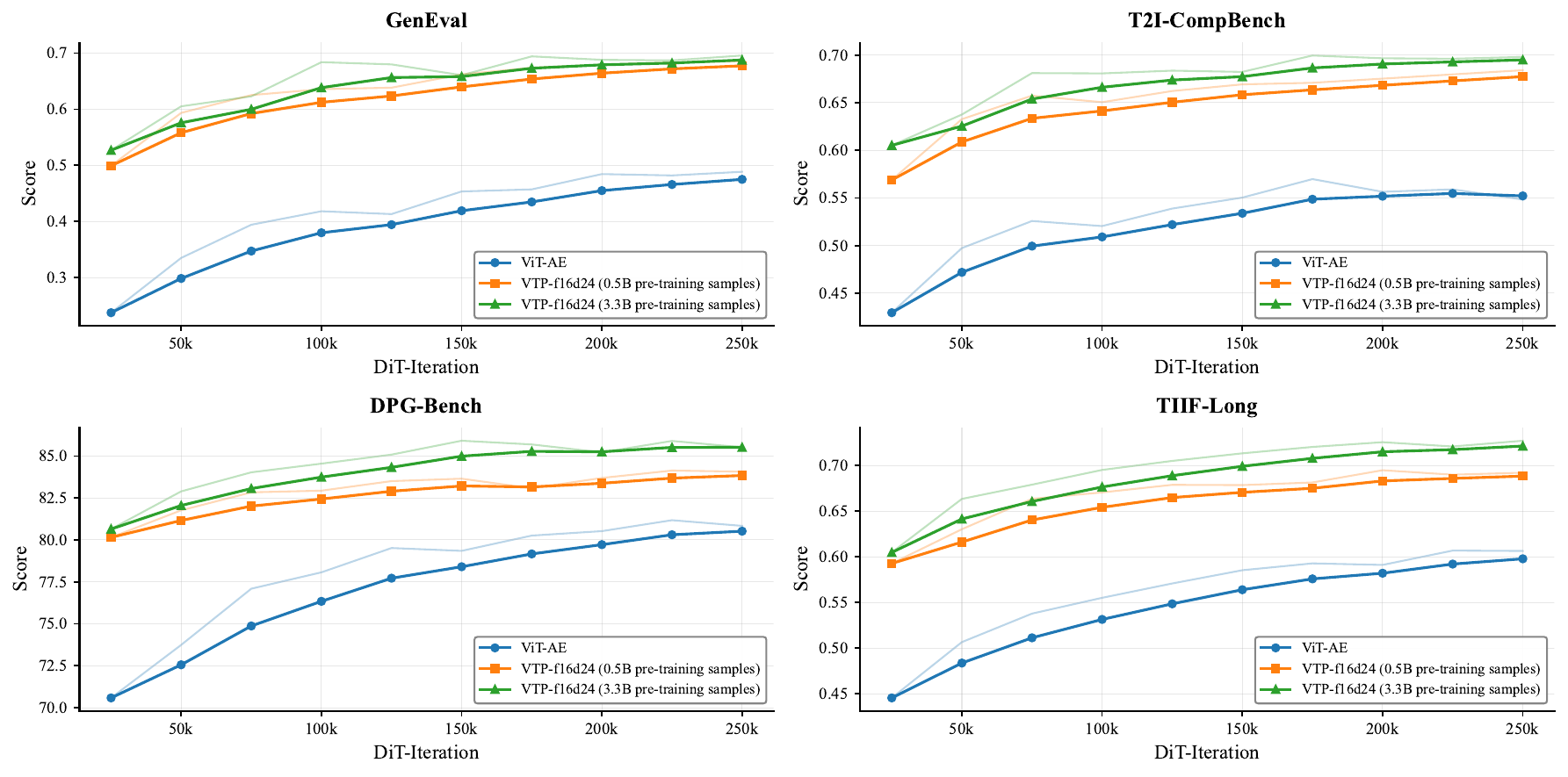}
        \caption{\textbf{VTP scaling properties generalize to T2I generation.} VTP pre-trained with representation learning objectives consistently improves with increased compute, achieving better generation performance than reconstruction-only tokenizers.}
        \label{fig:t2i_scaling}
    \end{subfigure}
    \hfill
    \begin{subfigure}[b]{0.34\textwidth}
        \centering
        \includegraphics[width=0.95\textwidth]{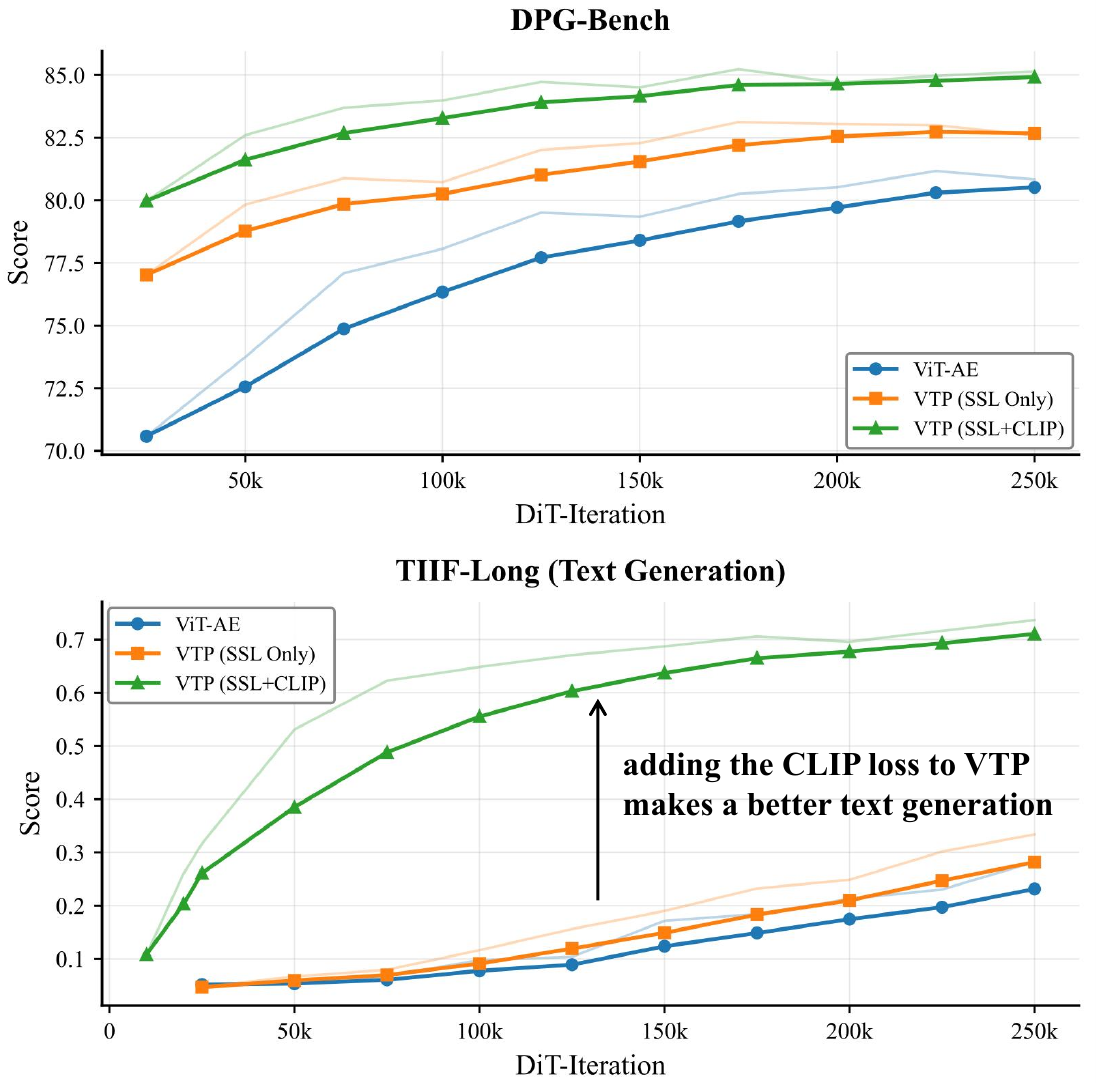}
        \caption{\textbf{Ablation on Semantic losses}. Adding CLIP loss to VTP significantly improves the text rendering ability in T2I.}
        \label{fig:t2i_ablation}
    \end{subfigure}
    \caption{\textbf{Text-to-Image Experiments on LAION.}}
    \label{fig:t2i_results}
\end{figure}

To validate the generalizability of VTP's scaling properties beyond class-conditional generation on ImageNet, we extend our evaluation to text-to-image (T2I) generation tasks on the LAION dataset~\cite{laion400m}.

We train VTP tokenizers and evaluate T2I generation performance across different pre-training compute budgets. For the downstream generative model, we adopt a DiT-XL architecture for generation. During training, images are resized with a short edge randomly sampled between 256 and 512. The results are shown in \cref{fig:t2i_scaling}. More details will be provided in the supplementary materials.

\paragraph{VTP scaling properties generalize to T2I.}
We observe two key findings:
(1) Tokenizer with representation learning objectives achieves significantly faster convergence than the reconstruction-only AE baseline, demonstrating that semantic-aware tokenizer pre-training substantially improves training efficiency for downstream T2I generation.
(2) As the tokenizer pre-training compute increases, the downstream generative model continues to improve, confirming that VTP's scaling properties are not limited to class-conditional ImageNet generation but generalize to the more challenging T2I setting.

\paragraph{CLIP loss improves text rendering.}
We further ablate the effect of different semantic losses on T2I generation quality (\cref{fig:t2i_ablation}). Similar to the results in \cref{fig:3in1} on ImageNet, as the variety of perception losses incorporated into VTP pre-training increases, we observe a steady improvement in downstream T2I generation performance. Furthermore, we observe that the tokenizer trained with the CLIP loss exhibits a significant advantage in text rendering for text-to-image generation, substantially outperforming both AE and SSL+AE and tokenizers. 

\section{Further Scaling and Comparison}
\label{sec:final_performance}

\begin{table}[t]
    \centering
    \small
    \renewcommand{\arraystretch}{1.1}
    \setlength{\tabcolsep}{4pt}
    \resizebox{\linewidth}{!}{%
    \begin{tabular}{l ccc c c cccc cccc}
    \toprule
    \multirow{2}{*}{\textbf{Method}} & \multicolumn{3}{c}{\textbf{Tokenizer}} & \multirow{2}{*}{\textbf{\makecell{Gen Model\\Params}}} & \multirow{2}{*}{\textbf{Epochs}} & \multicolumn{4}{c}{\textbf{Generation w/o guidance}} & \multicolumn{4}{c}{\textbf{Generation w/ guidance}} \\
    \cmidrule(lr){2-4} \cmidrule(lr){7-10} \cmidrule(lr){11-14}
     & \textbf{rFID}$\downarrow$ & \textbf{\makecell{Zero-\\Shot}}$\uparrow$ & \textbf{\makecell{Linear\\Probe}}$\uparrow$ & & & \textbf{gFID}$\downarrow$ & \textbf{IS}$\uparrow$ & \textbf{Prec.}$\uparrow$ & \textbf{Rec.}$\uparrow$ & \textbf{gFID}$\downarrow$ & \textbf{IS}$\uparrow$ & \textbf{Prec.}$\uparrow$ & \textbf{Rec.}$\uparrow$ \\
    \midrule
    \multicolumn{14}{c}{\textit{\textbf{Perception or Unified Tokenizer Baselines}}} \\
    \midrule
    SigLIP~\cite{siglip} & - & 80.5 & - & - & - & - & - & - & - & - & - & - & - \\
    MAE~\cite{mae} & - & - & 85.9 & - & - & - & - & - & - & - & - & - & - \\
    DINOv2~\cite{dinov2} & - & - & 86.7 & - & - & - & - & - & - & - & - & - & - \\
    VILA-U~\cite{vila-u} & 1.80 & 73.3 & - & - & - & - & - & - & - & - & - & - & - \\
    UniTok~\cite{unitok} & 0.41 & 70.8 & - & 1.4B & - & 2.51 & 216.7 & 0.82 & 0.57 & 2.77 & 227.5 & 0.81 & 0.57 \\
    \midrule
    \multicolumn{14}{c}{\textit{\textbf{Convergence Efficiency}}} \\
    \midrule
    REPA~\cite{repa}     & 0.61 & - & - & 675M & 80 & 7.90 & 122.6 & 0.70 & 0.65 & -    & -     & -    & -    \\
    DDT~\cite{ddt}      & 0.61 & - & - & 675M & 80 & 6.62 & 135.2 & 0.69 & \textbf{0.67} & 1.52 & 263.7 & 0.78 & 0.63 \\
    VA-VAE~\cite{vavae}   & \textbf{0.28} & - & - & 675M & 80 & 4.29 & -     & -    & -    & -    & -     & -    & -    \\
    REPA-E~\cite{repa-e}   & \textbf{0.28} & - & - & 675M & 80 & 3.46 & 159.8 & 0.77 & 0.63 & 1.67 & \textbf{266.3} & \textbf{0.80} & \textbf{0.63} \\
    RAE~\cite{rae}      & 0.57 & - & 84.5 & 675M & 80 & 4.28 & -     & -    & -    & -    & -     & -    & -    \\
    \textcolor{gray}{RAE~\cite{rae}} & \textcolor{gray}{0.57} & \textcolor{gray}{-} & \textcolor{gray}{84.5} & \textcolor{gray}{835M} & 80 & \textcolor{gray}{2.16} & \textcolor{gray}{214.8} & \textcolor{gray}{0.82} & \textcolor{gray}{0.59} & \textcolor{gray}{-} & \textcolor{gray}{-} & \textcolor{gray}{-} & \textcolor{gray}{-} \\
    \arrayrulecolor{gray!40}\midrule\arrayrulecolor{black} 
    \rowcolor{blue!8} \textbf{VTP (Ours)}      & 0.36 & \textbf{78.2} & \textbf{85.7} & 675M & 80 & \textbf{2.62} & \textbf{197.8} & \textbf{0.79}    & 0.62    & \textbf{1.44}    & 238.2     & \textbf{0.80}   & \textbf{0.63}    \\
    \rowcolor{blue!8} \textcolor{gray}{\textbf{VTP (Ours)}} & \textcolor{gray}{0.36} & \textcolor{gray}{78.2} & \textcolor{gray}{85.7} & \textcolor{gray}{1.0B} & 80 & \textcolor{gray}{2.03} & \textcolor{gray}{219.4} & \textcolor{gray}{0.80} & \textcolor{gray}{0.62} & \textcolor{gray}{-} & \textcolor{gray}{-} & \textcolor{gray}{-} & \textcolor{gray}{-} \\
    \midrule
    \multicolumn{14}{c}{\textit{\textbf{Long Period Training}}} \\
    \midrule
    DiT~\cite{dit}       & - & - & - & 675M & 1400 & 9.62 & 121.5 & 0.67 & 0.67 & 2.27 & 278.2 & \textbf{0.83} & 0.57 \\
    SiT~\cite{ma2024sit}       & - & - & - & 675M & 1400 & 8.61 & 131.7 & 0.68 & 0.67 & 2.06 & 270.3 & 0.82 & 0.59 \\
    VA-VAE~\cite{vavae}  & \textbf{0.28} & - & - & 675M & 800 & 2.17 & 205.6 & 0.77 & 0.65 & 1.35 & 295.3 & 0.79 & 0.65 \\
    REPA~\cite{repa}     & 0.61 & - & - & 675M & 800 & 5.78 & 158.3 & 0.70 & 0.68 & 1.29 & 306.3 & 0.79 & 0.64 \\
    DDT~\cite{ddt}       & 0.61 & - & - & 675M & 400 & 6.27 & 154.7 & 0.68 & \textbf{0.69} & 1.26 & \textbf{310.6} & 0.79 & 0.65 \\
    REPA-E~\cite{repa-e} & \textbf{0.28} & - & - & 675M & 800 & \textbf{1.70} & 217.3 & 0.77 & 0.66 & 1.15 & 304.0 & 0.79 & 0.66 \\
    RAE~\cite{rae}       & 0.57 & - & 84.5 & 676M & 800 & 1.87 & 209.7 & \textbf{0.80} & 0.63 & 1.41 & 309.4 & 0.80 & 0.63 \\
    \textcolor{gray}{RAE$^*$~\cite{rae}} & \textcolor{gray}{0.57} & \textcolor{gray}{-} & \textcolor{gray}{84.5} & \textcolor{gray}{839M} & \textcolor{gray}{800} & \textcolor{gray}{1.51} & \textcolor{gray}{242.9} & \textcolor{gray}{0.79} & \textcolor{gray}{0.63} & \textcolor{gray}{1.13} & \textcolor{gray}{262.6} & \textcolor{gray}{0.78} & \textcolor{gray}{0.67} \\
    \arrayrulecolor{gray!40}\midrule\arrayrulecolor{black} 
    \rowcolor{blue!8} \textbf{VTP (Ours)} & 0.36 & \textbf{78.2} & \textbf{85.7} & 675M & 600 & 1.85 & \textbf{232.3} & 0.79 & 0.63 & \textbf{1.11} & 279.5 & 0.79 & \textbf{0.67} \\
    \midrule
    \end{tabular}}
    \caption{\textbf{Generation Performance on ImageNet 256$\times$256.}}
    \label{tab:comparison_perf}
\end{table}

\begin{figure}[t]
    \centering
    \begin{subfigure}[b]{0.57\textwidth}
        \centering
        \includegraphics[width=\textwidth]{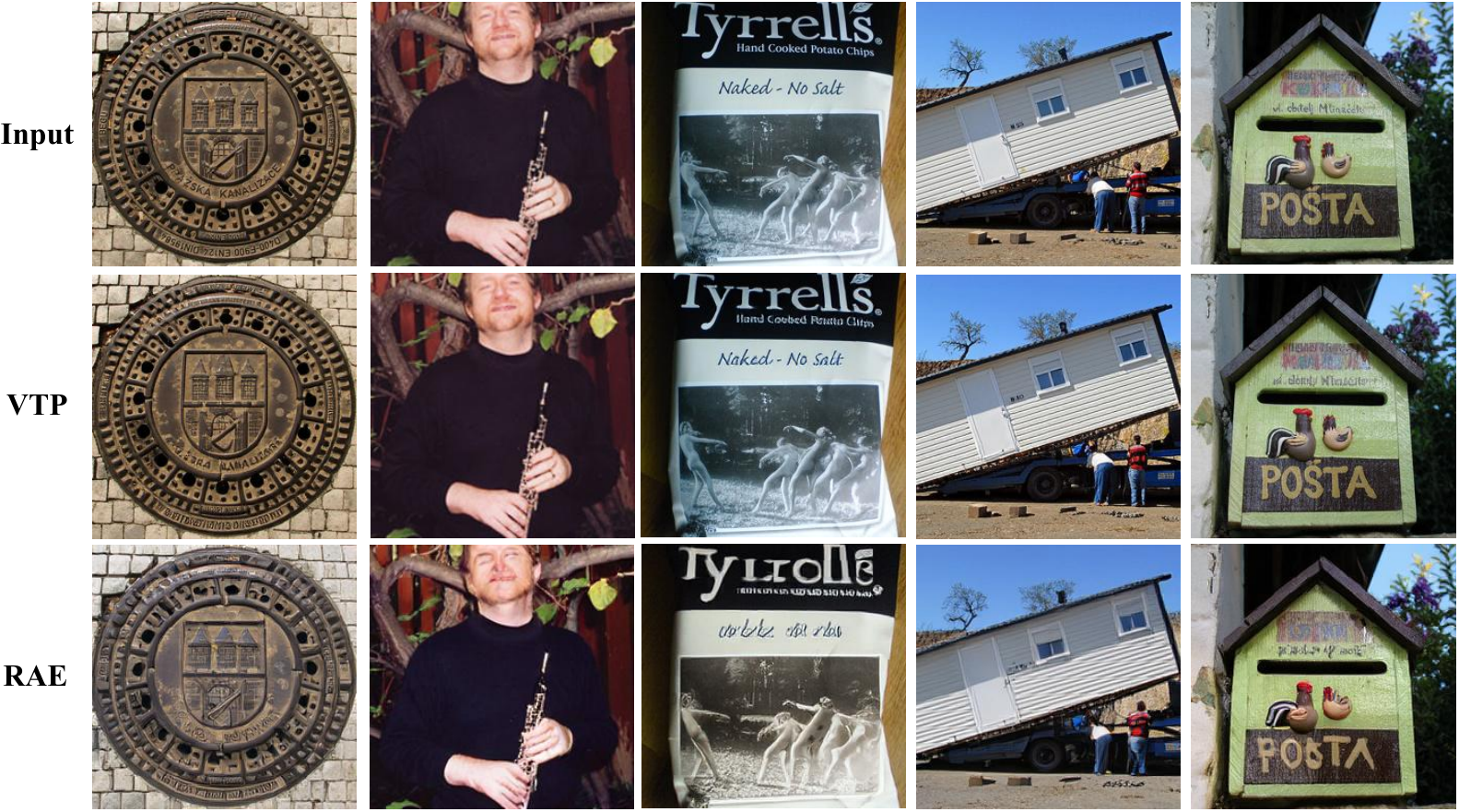}
        \caption{\textbf{Reconstruction Comparison with RAE.} Visual tokenizer pre-training with reconstruction target enables better reconstruction capacity than direct representation encoder transfer.}
        \label{fig:rae_comparing}
    \end{subfigure}
    \hfill
    \begin{subfigure}[b]{0.42\textwidth}
        \centering
        \includegraphics[width=\textwidth]{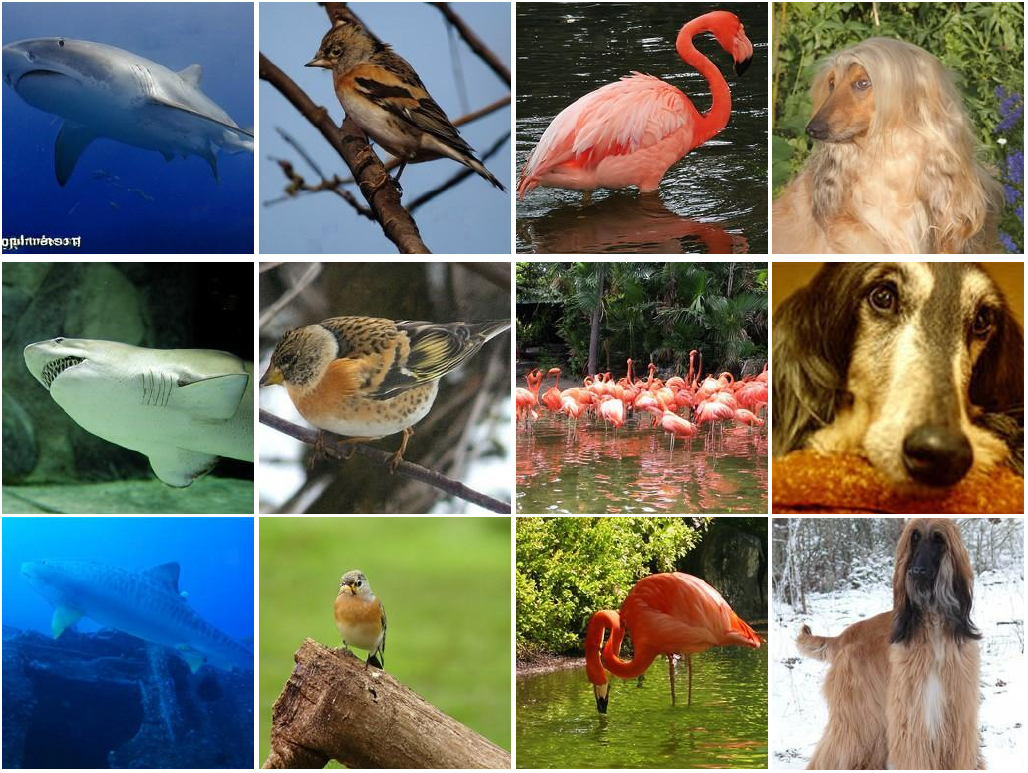}
        \caption{\textbf{Convergence on ImageNet.} With 80 epochs of training, VTP achieves 2.61~gFID, outperforming VA-VAE (4.29) and RAE (4.28).}
        \label{fig:comparing_vavae}
    \end{subfigure}
    \caption{\textbf{Visualizations of Reconstruction and Generation.}}
    \label{fig:comparison}
\end{figure}

We further scale VTP pre-training with larger training compute across three symmetric ViT architectures (VTP-S, VTP-B, VTP-L) and summarize results in \cref{tab:scale_rae} and \cref{tab:comparison_perf}.

\paragraph{Scalability Compared to Fixed Representation AutoEncoders.}
Compared to concurrent work RAE~\cite{rae}.
A distinctive advantage of VTP is its scalability. As shown in Table~\ref{tab:scale_rae}, under identical DiT training configurations, VTP's downstream generation improves consistently as the tokenizer scales from S to L, whereas RAE's performance degrades at larger scales. Moreover, since VTP always involves reconstruction during training, it preserves fine-grained details significantly better than RAE (see \cref{fig:rae_comparing}).

\begin{wraptable}[8]{r}{0.42\textwidth}
    \vspace{-24pt}
    \centering
    \small
    \setlength{\tabcolsep}{6pt}
    \begin{tabular}{l cc}
    \toprule
    \textbf{Tokenizer} & \textbf{RAE} & \textbf{VTP} \\
    \midrule
    Small & 3.50 & 5.46 \\
    Base & 4.28 & 3.88 \\
    Large & 6.09 & \textbf{2.81} \\
    \bottomrule
    \end{tabular}
    \caption{\textbf{Scalability comparison.} LightningDiT gFID$\downarrow$ at 80 epochs (w/o guidance) with identical DiT training. We take RAE's results from its original paper.}
    \label{tab:scale_rae}
    \vspace{-36pt}
\end{wraptable}
\paragraph{Unified Performance Frontier.}
Our final model achieves 0.36 rFID, 78.2\% zero-shot accuracy, and 85.7\% linear probing accuracy on ImageNet, surpassing prior unified tokenizers VILA-U~\cite{vila-u} and UniTok~\cite{unitok}.

\paragraph{Superior Generation Without Architecture Modification.}
Built upon the standard LightningDiT~\cite{vavae}, VTP-L achieves superior 1.11 gFID with guidance, surpassing all prior methods (\cref{tab:comparison_perf}). Moreover, VTP attains 2.60 and 2.03 gFID without guidance in only 80 epochs, demonstrating remarkably fast convergence. More details will be provided in supplementary materials.

\section{Conclusion}

In this work, we redesign a scalable paradigm for visual tokenizer pre-training that substantially enhances generative performance and propose \modelname. We demonstrate that perception-oriented tokenizer pre-training unlocks a new scaling law for generation: (1) semantic understanding is the key driver for improving generation, and (2) with this understanding, the visual tokenizer achieves scalable performance on generative tasks. In contrast to traditional tokenizers pre-trained solely on reconstruction—whose performance saturates with a small scale—our approach consistent attains a significant gain in generative performance when scaling the compute budget, model size, data scales. Hope VTP could inpire the following research on visual tokenizers.


\bibliography{main}
\end{document}